\documentclass[sigconf]{acmart}

\usepackage{hyperref}
\renewcommand\footnotetextcopyrightpermission[1]{}

\AtBeginDocument{%
  }

\setcopyright{acmlicensed}
\copyrightyear{2018}
\acmYear{2018}
\acmDOI{XXXXXXX.XXXXXXX}

\acmConference[Conference acronym 'XX]{Make sure to enter the correct
  conference title from your rights confirmation email}{June 03--05,
  2018}{Woodstock, NY}

\acmISBN{978-1-4503-XXXX-X/2018/06}

\begin{document}

\title{\raisebox{-1.5ex}{\includegraphics[height=1.8em]{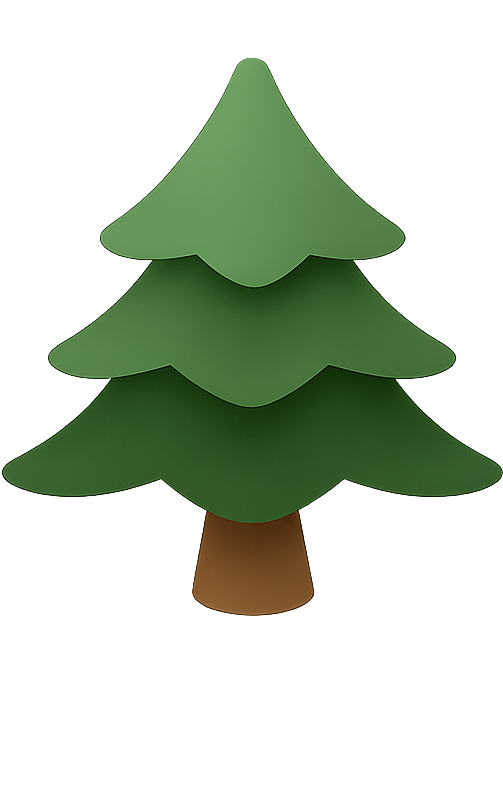}}\hspace{0.2em}\normalfont\bfseries AgriGPT-Omni: A Unified Speech–Vision–Text Framework for Multilingual Agricultural Intelligence}

\renewcommand{\shortauthors}{Trovato et al.}

\author{Bo Yang}
\affiliation{%
  \institution{Zhejiang University}
  \city{Hangzhou}
  \country{China}}
\email{boyang30@zju.edu.cn}

\author{Lanfei Feng}
\affiliation{%
  \institution{Zhejiang University}
  \city{Ningbo}
  \country{China}}
\email{22451116@zju.edu.cn}

\author{Yunkui Chen}
\affiliation{%
  \institution{Zhejiang University}
  \city{Ningbo}
  \country{China}}
\email{22351048@zju.edu.cn}

\author{Yu Zhang}
\affiliation{%
  \institution{Zhejiang University}
  \city{Hangzhou}
  \country{China}}
\email{22421173@zju.edu.cn}

\author{Jianyu Zhang}
\affiliation{%
  \institution{Zhejiang University}
  \city{Hangzhou}
  \country{China}}
\email{jianyu.zhang@zju.edu.cn}

\author{Xiao Xu}
\affiliation{%
  \institution{Zhejiang University}
  \city{Hangzhou}
  \country{China}}
\email{3200105334@zju.edu.cn}

\author{Nueraili Aierken}
\affiliation{%
  \institution{Zhejiang University}
  \city{Hangzhou}
  \country{China}}
\email{nureli@zju.edu.cn}

\author{Shijian Li\textsuperscript{*}}
\affiliation{%
  \institution{Zhejiang University}
  \city{Hangzhou}
  \country{China}}
\email{shijianli@zju.edu.cn}

\begin{abstract}
  Despite rapid advances in multimodal large language models, agricultural applications remain constrained by the lack of multilingual speech data, unified multimodal architectures, and comprehensive evaluation benchmarks. To address these challenges, we present AgriGPT-Omni, an agricultural omni-framework that integrates speech, vision, and text in a unified framework.(1) First, we construct a scalable data synthesis and collection pipeline that converts agricultural texts and images into training data, resulting in the largest agricultural speech dataset to date, including 492K synthetic and 1.4K real speech samples across six languages.(2) Second, based on this, we train the first agricultural Omni-model via a three-stage paradigm: textual knowledge injection, progressive multimodal alignment, and GRPO-based reinforcement learning, enabling unified reasoning across languages and modalities.(3) We further propose AgriBench-Omni-2K, the first tri-modal benchmark for agriculture, covering diverse speech–vision–text tasks and multilingual slices, with standardized protocols and reproducible tools. Experiments show that AgriGPT-Omni significantly outperforms general-purpose baselines on multilingual and multimodal reasoning as well as real-world speech understanding. All models, data, benchmarks, and code will be released to promote reproducible research, inclusive agricultural intelligence, and sustainable AI development for low-resource regions.
\end{abstract}


\received{20 February 2007}
\received[revised]{12 March 2009}
\received[accepted]{5 June 2009}

\maketitle

\begin{figure*}[htbp]
  \centering
  \includegraphics[width=\textwidth]{v3_train_overview-10.png}
  \vspace{0.1pt}  
  \caption{Three-stage training pipeline of AgriGPT-Omni. AgriGPT-Omni is trained through (1) text knowledge injection, including 2.2B-domain-token continued pretraining and 342K text/audio–text QA SFT; (2) multimodal knowledge learning, using 600K vision–language pairs, 342K audio–text pairs, and progressively unfreezing on 800K VQA, 50K high-quality VQA, and 140K audio-VQA; and (3) GRPO reinforcement learning with 2.5K audio-QA, 5K audio-VQA, and 1.4K human-recorded transcription pairs. The full pipeline covers six languages and yields the final AgriGPT-Omni checkpoint.}
  \label{fig:overview training}
\end{figure*}

\section{Introduction}

With the rapid progress of large language models (LLMs) and multimodal large language models (MLLMs), unified language–image–speech modeling has made significant strides across general domains \citep{brown2020language,hurst2024gpt,alayrac2022flamingo,lu2023ziyavisual,gemini2023,bai2023qwenvl}. Recent omni-modal systems such as GPT-4o and Gemini 2.0 demonstrate increasingly fluid integration of perception, reasoning, and generation. In agriculture, emerging multimodal efforts—e.g., AgriGPT-VL \citep{yang2025agrigptvl}, AgriDoctor \citep{zhang2025agridoc}, and AgriBench \citep{zhou2024agribench}—begin exploring domain-specific vision–language modeling, yet remain largely text–image oriented without unified speech integration. Consequently, most progress remains concentrated in open-domain applications, while agriculture has yet to benefit from full-spectrum language–vision–speech unification.

Agriculture is inherently multimodal and context-dependent. Real-world field scenarios require accurate speech understanding, robust visual perception, and precise text-based agronomic reasoning. Prior agricultural multimodal systems—including VL frameworks \citep{yu2024agriVLframework} and diagnostic models for crop stresses \citep{long2025multimodaldiag}—primarily address image-centric or structured QA tasks. However, existing agricultural AI research remains heavily skewed toward vision–language QA, exemplified by datasets such as PlantVillage, IP102, and recent text–image systems like AgroGPT \citep{hughes2015plantvillage,wu2019ip102,awais2025agrogpt}. The speech modality—arguably the most natural interface for farmers—remains severely under-resourced, and domain LLMs such as AgriBERT, AgriGPT, AgroLLM, and AgriLLM \citep{rezayi2022agribert,yang2025agrigpt,samuel2025agrollm,didwania2024agrillm} lack unified tri-modal capabilities. This absence of a speech-centric tri-modal benchmark further limits meaningful cross-modal evaluation.

To address these gaps, we introduce \textbf{AgriGPT-Omni}, the first complete system in agriculture that supports unified modeling and reasoning over speech, images, and text. We systematically construct the framework around the three pillars of \emph{data, model, and evaluation}.  
\begin{itemize}
  \item \textbf{Speech dataset:} We construct the first agricultural speech corpus via a hybrid synthetic–human pipeline, totaling \textbf{492K} synthetic utterances and \textbf{1.4K} human recordings across six languages , covering multiple task formats and validated in real application scenarios.

  \item \textbf{Omni-modal model:} We train \textbf{AgriGPT-Omni}, a unified architecture integrating speech, image, and text inputs, with stepwise alignment and a GRPO-based reward objective to enhance tri-modal understanding and output stability. 
  
  \item \textbf{Omni-modal benchmark:} We build the first fully multimodal agricultural benchmark covering composite tasks in six languages, with standardized evaluation protocols and toolchains for consistent assessment and comparison.

All models, datasets, and evaluation tools will be released to enable agriculture to move from ``text--image understanding'' toward \emph{true} ``omni-modal interaction.'' 

\end{itemize}

\begin{figure*}[htbp]
  \centering
  \includegraphics[width=\textwidth]{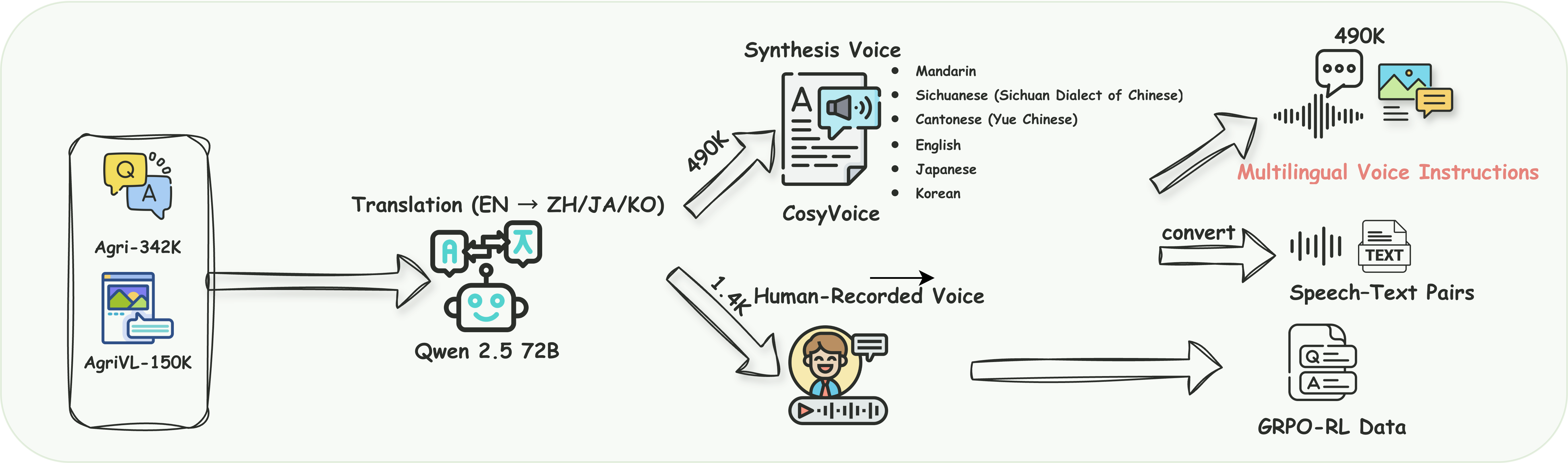}
  \caption{We build AgriGPT-Omni’s multilingual speech data by translating Agri-342K QA and AgriVL-150K image–text samples into six languages using Qwen2.5-72B, then synthesizing ~490K speech clips with CosyVoice-0.5B. We further record ~1.4K human utterances (filtered to 1,431 for training) and collect 600 real-world evaluation recordings (586 retained after review). All audio is converted into speech–text pairs, with 1,431 real dictation samples used for GRPO.}
  \label{fig:data construction}
\end{figure*}

\section{Related Work}
\subsection{General and agricultural large models}

General-purpose MLLMs have rapidly unified language–image–speech, exemplified by GPT-4, Gemini, Qwen2.5-Omni, Kosmos-1, and SeamlessM4T, and now show strong open-domain generation and reasoning with large cross-modal corpora and alignment strategies \citep{hurst2024gpt,gemini2023,xu2025qwen25omni,huang2023kosmos1,barrault2023seamlessm4t}. 
In agriculture, progress moved from text-centric models (AgriBERT; AgriLLM; AgroLLM) to vision–language systems (AgroGPT; AgriGPT with Tri-RAG), yet most remain single or bi-modal, lacking unified speech–image–text modeling and standardized omni-modal evaluation for real-world voice-centric workflows \citep{rezayi2022agribert,didwania2024agrillm,samuel2025agrollm,awais2025agrogpt,yang2025agrigpt}.

\subsection{Omni data resources}

Text resources now standardize instruction-style agricultural knowledge: Agri-342K with AgriBench-13K/Mini-AgriBench600 in AgriGPT, large-scale real farmer queries in AgriLLM, and AgroInstruct from AgroGPT \citep{yang2025agrigpt,didwania2024agrillm,awais2025agrogpt}. 
For vision–language, benchmarks capture expert knowledge and fine-grained categories, including AgMMU, VL-PAW, and AgroBench \citep{gauba2025agmmu,yu2025vlpaw,shinoda2025agrobench}. 
The missing piece is speech: despite general multilingual corpora (Common Voice, SLUE/SLUE-Phase2, FLEURS, VoxPopuli, MLS, MMS) \citep{ardila2020commonvoice,shon2021slue,shon2023sluephase2,conneau2022fleurs,wang2021voxpopuli,pratap2020mls,pratap2023mms}, there is no agriculture-specific speech dataset—especially for speech QA, speech+image understanding, and multilingual agricultural instructions. We therefore introduce a unified speech data pipeline for six languages (QA, multiple choice, transcription) with human recordings for validation.

\subsection{Omni benchmarks and evaluation}

Standardized benchmarks underpin comparability and reproducibility. For vision–language, VQA v2, GQA, ScienceQA, MMBench, MMMU, and SEED-Bench span diverse formats and reasoning levels \citep{goyal2017vqav2,hudson2019gqa,lu2022scienceqa,liu2023mmbench,yue2024mmmu,li2023seedbench}. 
For speech and speech–text, SUPERB and SLUE (including Phase-2) provide multi-task SLU/ASR suites, while MMS scales multilingual speech technology to over a thousand languages \citep{yang2021superb,shon2023sluephase2,pratap2023mms}. 
However, agriculture lacks a unified benchmark with speech input and speech–image reasoning as well as consistent judging protocols. We release an omni-modal suite with four tasks (speech QA, speech multiple choice, speech–image QA, speech–image multiple choice) and six languages, together with symmetric evaluation and rubric-guided judging scripts to reduce position bias and format variance \citep{zheng2023mtbench,hashemi2024llmrubric}.

\section{AgriGPT-Omni}

\subsection{Dataset Construction}

As shown in Figure~\ref{fig:data construction}, we build the first large-scale multilingual speech-vision-text dataset for agriculture, combining synthesized and real speech. The dataset consists of three main components:

\begin{itemize}
    \item \textbf{Synthesized speech:} We convert the multilingual version of Agri-342K (covering Chinese, Sichuan dialect, Cantonese, English, Japanese, and Korean) into speech using the \textbf{CosyVoice2-0.5B} TTS system, yielding about \textbf{342K} speech–text pairs. In addition, we sample \textbf{150K} image-grounded QA pairs from Agri-VL-3M and also synthesize them into speech. In total, we produce about \textbf{492K} high-quality synthesized speech samples covering QA and multiple-choice tasks.
    
    \item \textbf{Real speech:} \textbf{In particular}, to enhance the model’s robustness to real-world conditions, we collect \textbf{2,200} real speech recordings from multilingual volunteers under controlled conditions across the same six languages. The recordings include \textbf{1,500} multiple-choice samples (used for training) and \textbf{586} transcription samples (used only for evaluation). These real speech samples mimic accent and noise variations and provide a realistic cross-modal evaluation baseline.
    
    \item \textbf{Reinforcement-learning subset:} From the above \textbf{492K} synthesized speech samples, we construct a carefully selected \textbf{8,931}-sample reward-optimization subset, including speech multiple-choice, speech–image multiple-choice, and speech transcription tasks. This subset has clear structures and gold-standard answers, enabling exact-match or edit-distance rewards for Group Relative Policy Optimization (GRPO) in the training stage.
\end{itemize}

This unified dataset covers speech, image, and text modalities with diverse languages and task types, providing a solid foundation for AgriGPT-Omni to achieve cross-modal and cross-lingual reasoning.

\begin{figure*}[htbp]
  \centering
  \includegraphics[width=\textwidth]{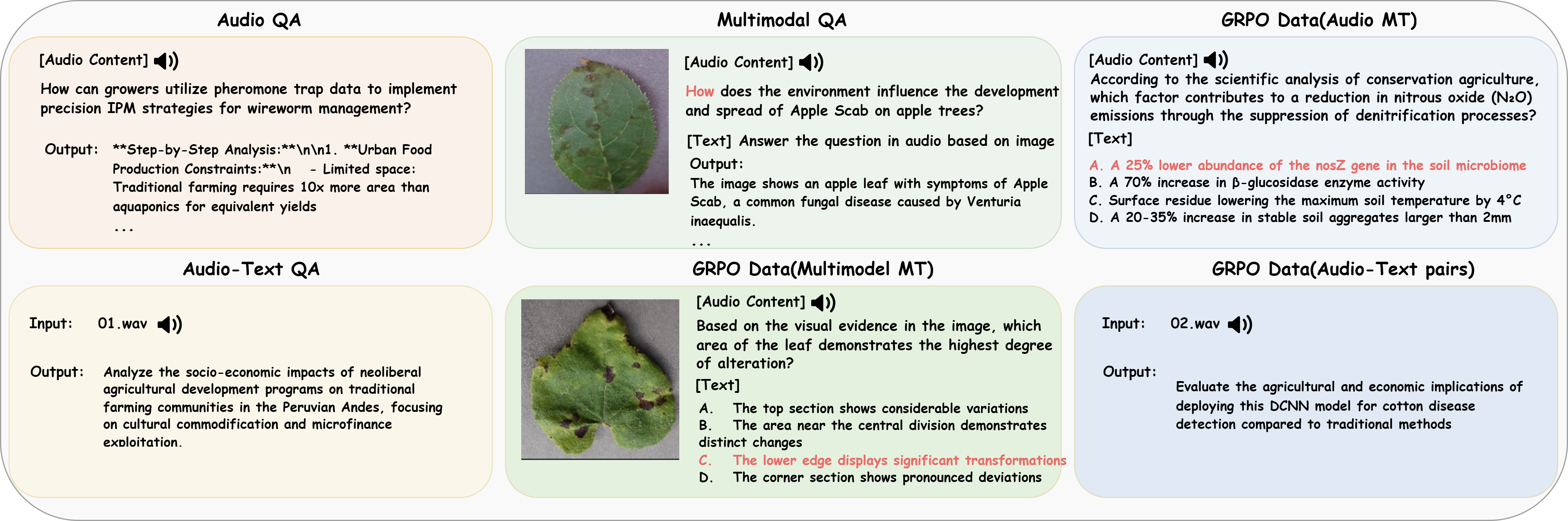}
  \caption{Example data types used in AgriGPT-Omni Training.
(1) Audio QA — answering questions from speech input;
(2) Multimodal QA — speech questions combined with image-based reasoning;
(3) GRPO Audio Multiple-Choice — selecting an answer from speech-only multiple-choice queries;
(4) Audio–Text QA — answering questions from audio files;
(5) GRPO Multimodal Multiple-Choice — speech + image inputs for multimodal MC tasks;
(6) GRPO Audio–Text Pairs — transcribing speech to text for ASR-oriented reinforcement learning.}
  \label{fig:training data show}
\end{figure*}

\subsection{Model training}

As shown in Figure~\ref{fig:overview training}, we build on \textsc{Qwen-2.5-Omni} and adopt a three-stage curriculum---\emph{text knowledge injection} $\rightarrow$ \emph{multimodal alignment \& instruction tuning} $\rightarrow$ \emph{preference optimization}---so that language competence and task formatting are first consolidated, cross-modal mappings are then learned along stable gradient paths, and final preference signals improve controllability and generalization stability.

\paragraph{Stage 1: Text knowledge injection.}
\textbf{(1) Continued pretraining.} We continue pretraining on agricultural corpora for \textbf{2.2B tokens}, expanding domain terminology coverage and compositional generalization to reduce the language-side bottleneck in later multimodal training. 
\textbf{(2) Pure-text instruction tuning.} We fine-tune on \textbf{342k} text-only QA instances to normalize instruction-following behavior and response style toward agricultural tasks by updating the \emph{language backbone}. 
\textbf{(3) Speech--text alignment tuning.} We train on \textbf{342k} ``spoken question $\rightarrow$ textual answer'' pairs \emph{without freezing}: the language backbone, the audio encoder, and the speech--language adapters are jointly updated to establish an end-to-end listen-to-answer pathway. 

\noindent After Stage~1, the model exhibits solid language modeling and instruction-following, with an established end-to-end speech QA pipeline; the language prior also anchors subsequent alignment steps and lowers the entry barrier for multimodal learning.

\paragraph{Stage 2: Multimodal alignment and instruction tuning.}
\textbf{(1) Vision--language alignment (frozen).} On \textbf{600K} image--caption pairs we freeze the vision encoder and the language backbone, updating only the vision--language adapters to obtain a stable mapping. 
\textbf{(2) Speech--language alignment (frozen).} On \textbf{342k} audio-alignment pairs we freeze the audio encoder and the language backbone, updating only the speech--language adapters so that acoustic representations align with the language space. 
\textbf{(3) Unfrozen instruction/VQA tuning.} After alignment, we unfreeze the language backbone and jointly optimize on \textbf{800K} multimodal instruction/VQA examples, coupling generation with multimodal representations. 
\textbf{(4) High-quality refinement.} We conduct a small-step update on \textbf{50k} carefully curated GPT-4o samples to correct long-tail errors and format inconsistencies from large-scale tuning, enhancing fine-grained attribute recognition and answer stability. 
\textbf{(5) Tri-modal QA tuning.} Finally, on \textbf{140k} ``speech + image $\rightarrow$ text'' instances we keep the language backbone unfrozen and jointly optimize the cross-modal adapters to strengthen tri-modal coordination. 

\noindent After Stage~2, robust vision/speech--language alignments are established; unfrozen joint tuning internalizes multimodal knowledge into language generation, while refinement and tri-modal QA improve detail sensitivity, cross-modal consistency, and response robustness, yielding a unified multimodal instruction model.

\paragraph{Stage 3: Preference Optimization (GRPO)}
We freeze the Stage~2 model as the \emph{reference policy} $\pi_{\mathrm{ref}}$ and optimize the trainable policy $\pi_{\phi}$ with GRPO~\citep{Shao2024DeepSeekMath}. Training uses three preference sources: \textbf{2{,}500} speech$\rightarrow$text multiple-choice items, \textbf{5{,}000} speech/image/text multiple-choice items, and \textbf{1{,}431} speech–transcription pairs. We jointly update the language backbone and cross-modal adapters, while the vision/audio encoders follow the Stage~2 setting. The learning objective adopts group-standardized advantages together with a PPO-style clipped surrogate and a KL penalty to the reference policy~\citep{Schulman2017PPO,Shao2024DeepSeekMath}; detailed are deferred to \textbf{Appendix~A.1}.

This stage consistently improves multiple-choice accuracy as measured by EM~\citep{Rajpurkar2016SQuAD}, reduces transcription errors in terms of WER/CER~\citep{JurafskyMartinSLP3,Levenshtein1966}, and yields more stable tri-modal QA under different sampling temperatures and random seeds. We also observe fewer option-leakage artifacts and more consistent formatting across tasks. Together, these gains complete the transition from an aligned model to a \emph{controllable and robust} \textsc{AgriGPT-Omni}.

To ensure full reproducibility, the specifications of our experimental hardware and the hyperparameters used for model training are documented in \textbf{Appendix~A.2}.

\begin{figure*}[htbp]
  \centering
  \includegraphics[width=\textwidth]{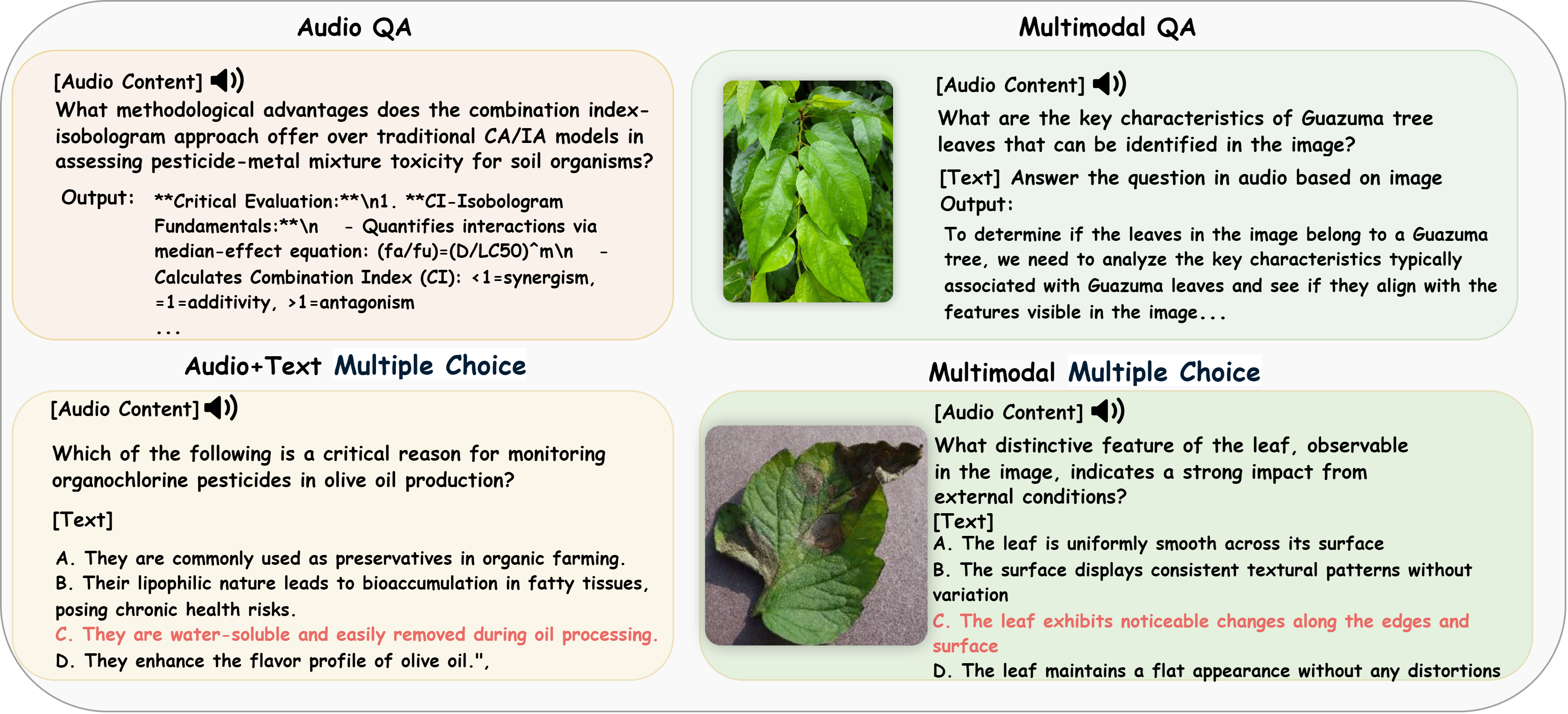}
  \caption{
Benchmark datasets cases show. 
(1) \textbf{Audio QA}: the model answers solely based on spoken queries; 
(2) \textbf{Audio--Image QA}: speech queries require grounding in visual evidence; 
(3) \textbf{Audio--Text Multiple Choice}: speech input combined with textual options; 
(4) \textbf{Audio--Image--Text Multiple Choice}: full tri-modal reasoning integrating audio queries, images, and textual choices. 
}
  \label{fig:bencamark data case}
\end{figure*}

\begin{figure*}[htbp]
  \centering
  \includegraphics[width=\textwidth]{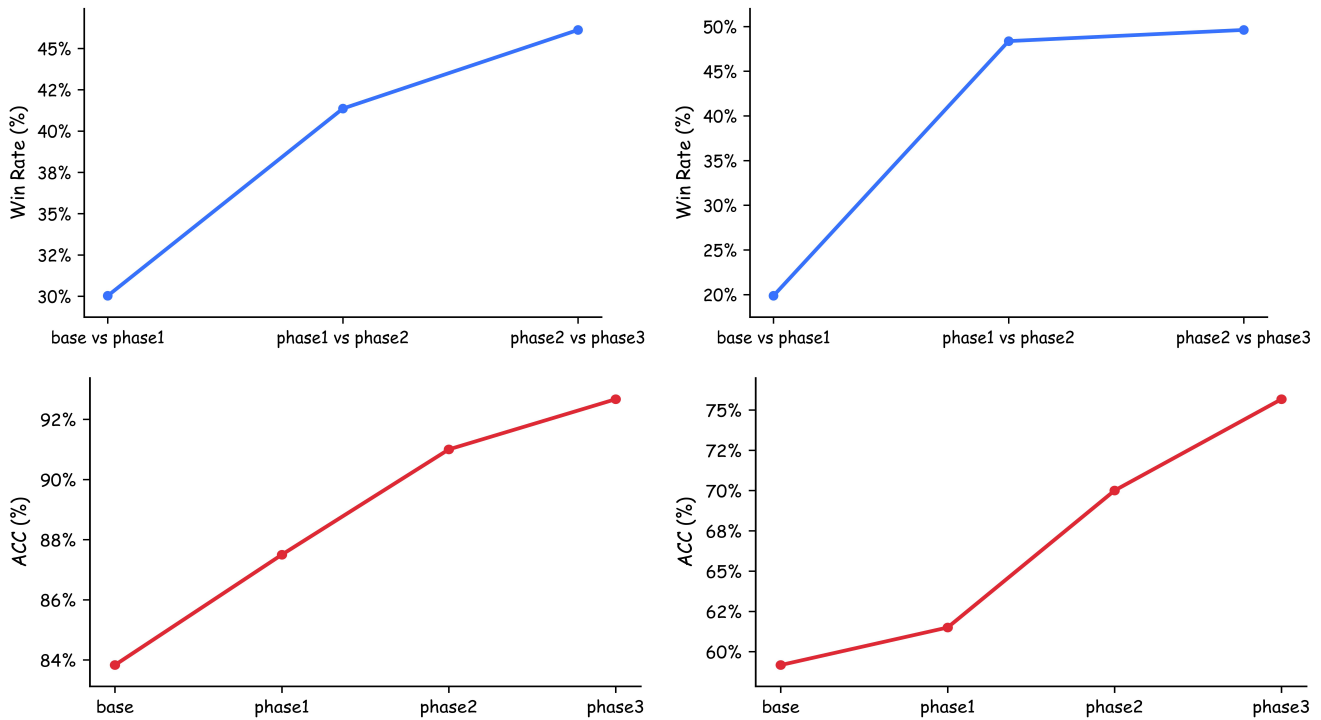}
  \caption{Ablation study results.
From left to right and top to bottom:
(1) Open QA (Speech → Text),
(2) Open VQA (Speech + Image → Text),
(3) Multiple Choice (Speech + Text → Text), and
(4) Multiple Choice (Speech + Image + Text → Text).
}
  \label{fig:ablation}
\end{figure*}

\begin{table*}[htbp]
\caption{\textbf{Text-only generation results on AgriBench-13K.} All scores are percentages. The best result in each column is \textbf{bold}, and the second-best is \underline{underlined}.}
\label{tab:agribench13k-text}
\centering
\small
\renewcommand{\arraystretch}{1.2}
\setlength{\tabcolsep}{20pt}
\begin{tabular}{lccccc}
\toprule
\textbf{Model} & \textbf{Bleu} & \textbf{Meteor} & \textbf{Rouge-1-f} & \textbf{Rouge-2-f} & \textbf{Rouge-l-f} \\
\midrule
InternVL-3-8B & 5.52 & 23.07 & 24.14 & 5.69 & \underline{23.07} \\
Llava-1.5-7B & 1.44 & 13.62 & 21.67 & 4.88 & 20.60 \\
MiniCPM-V-2.6-8B & 1.15 & 12.50 & 21.14 & 5.25 & 20.15 \\
Yi-VL-6B & 1.31 & 13.28 & 20.29 & 4.36 & 19.28 \\
Yi-VL-34B & 1.68 & 14.27 & 21.28 & 4.74 & 20.27 \\
Qwen2.5-VL-7B-Instruct & \underline{7.69} & \underline{30.17} & 24.16 & 4.97 & 22.86 \\
QVQ-72B-Preview & 2.48 & 17.31 & 17.54 & 3.63 & 16.75 \\
Deepseek-VL2 & 6.37 & 29.67 & 22.05 & \underline{4.93} & 20.93 \\
Gemini-2.5-Flash & 6.12 & 27.49 & \underline{24.85} & \textbf{5.59} & 23.73 \\
Gemini-2.5-Pro & 4.35 & 24.39 & 22.45 & 4.40 & 21.47 \\
\textbf{AgriGPT-Omni(Ours-8b)} & \textbf{9.69} & \textbf{30.37} & \textbf{26.88} & 5.49 & \textbf{25.53} \\
\bottomrule
\end{tabular}

\end{table*}


\begin{table*}[htbp]
\caption{\textbf{Image-text generation results on AgriBench-VL-4K.} All scores are percentages. The best result in each column is \textbf{bold}, and the second-best is \underline{underlined}.}
\label{tab:agribench-vl4k}
\centering
\small
\renewcommand{\arraystretch}{1.2}
\setlength{\tabcolsep}{20pt}
\begin{tabular}{lccccc}
\toprule
\textbf{Model} & \textbf{Bleu} & \textbf{Meteor} & \textbf{Rouge-1-f} & \textbf{Rouge-2-f} & \textbf{Rouge-l-f} \\
\midrule
InternVL-3-8B & 5.38 & 21.85 & 33.43 & 12.22 & 30.66 \\
Llava-1.5-7B & 2.39 & 15.55 & 30.96 & 11.09 & 28.56 \\
MiniCPM-V-2.6-8B & 7.16 & 22.57 & \underline{36.01} & \underline{13.39} & \underline{33.36} \\
Yi-VL-6B & 2.21 & 15.21 & 30.33 & 10.59 & 27.88 \\
Yi-VL-34B & 2.82 & 16.60 & 32.06 & 11.07 & 29.41 \\
Qwen2.5-VL-7B-Instruct & \underline{13.42} & \textbf{38.24} & 35.52 & 10.78 & 32.73 \\
QVQ-72B-Preview & 4.55 & 19.43 & 31.49 & 9.98 & 29.53 \\
Deepseek-VL2 & 2.89 & 30.85 & 25.57 & 7.12 & 22.98 \\
Gemini-2.5-Flash & 9.16 & 29.38 & 33.49 & 9.97 & 31.03 \\
Gemini-2.5-Pro & 6.55 & 26.22 & 30.25 & 8.64 & 28.00 \\
\textbf{AgriGPT-Omni(Ours-8b)} & \textbf{16.89} & \underline{37.72} & \textbf{42.25} & \textbf{14.62} & \textbf{39.41} \\
\bottomrule
\end{tabular}

\end{table*}

\begin{table}[htbp]
\centering
\caption{Pairwise Comparison Between Synthetic Speech and Human Speech}
\label{tab:synthetic-vs-human}
\small
\setlength{\tabcolsep}{10pt}
\begin{tabular}{lccc}
\toprule
Comparison & Win & Tie & Loss \\
\midrule
Synthetic Speech vs Human Speech & 234 & 111 & 232 \\
\bottomrule
\end{tabular}
\end{table}

\subsection{AgriBench-Omni-2K: A Full-Modality Evaluation Benchmark}

To comprehensively evaluate model performance in agricultural multimodal scenarios, we construct \textbf{AgriBench-Omni}, the first benchmark covering speech, image, and text modalities across multiple task formats.

We design four representative task types:

\begin{itemize}
  \item \textbf{Audio QA}: The input is a spoken question. The model must generate a free-form text answer, testing speech comprehension and generation capabilities.
  \item \textbf{Audio+Text Multiple Choice}: Given a spoken question and several textual choices, the model selects the correct answer, assessing speech parsing and semantic understanding.
  \item \textbf{Multimodal QA}: The model receives a spoken question and an image, and must generate a descriptive answer based on both modalities, evaluating multimodal perception and reasoning.
  \item \textbf{Multimodal Multiple Choice}: The input consists of a spoken question, an image, and textual choices. The model selects the correct answer, testing tri-modal alignment and decision-making.
\end{itemize}

We construct a total of \textbf{1,500 evaluation samples} across these tasks, composed of \textbf{100 questions} $\times$ \textbf{6 languages} $\times$ \textbf{4 task types}. The supported languages include Chinese, Sichuan dialect, Cantonese, English, Japanese, and Korean. \textbf{All samples are manually reviewed} by human annotators to ensure quality and linguistic correctness.

To test real-world audio robustness, we further collected \textbf{600 human-recorded speech samples} (100 per language), and filtered out 14 low-quality samples through manual review, resulting in a final set of \textbf{586 high-quality real speech recordings} covering all six languages and task types.

To ensure zero overlap between training and evaluation data, we applied a \textbf{dual de-duplication process}: (1) computing \textbf{ROUGE-L similarity} between benchmark and training texts, discarding any sample with a similarity score above 0.7; (2) using GPT-4-based semantic filtering to identify and remove deeper duplications not captured by surface metrics.

Each sample is validated by domain experts and follows a unified scoring protocol with reproducible evaluation scripts. AgriBench-Omni fills a critical gap in agricultural multimodal benchmarks and provides a standard platform for rigorous comparison, alignment, and deployment of domain-specific models.

\section{Result}
\subsection{Comparative Experiments} 

 We benchmark AgriGPT-Omni against ten representative multimodal models that cover open-source and production systems: InternVL-3-8B \cite{internvl3_2025,internvl3_blog_2025}, LLaVA-1.5-7B \cite{llava15}, MiniCPM-V-2.6-8B \cite{minicpmv26_arxiv}, Yi-VL-6B/34B \cite{yi_arxiv}, Qwen2.5-VL-7B-Instruct \cite{qwen25_arxiv}, Qwen-QVQ-72B-Preview \cite{qvq72b_blog}, DeepSeek-VL2 \cite{deepseek_vl2}, Google Gemini 2.5 Flash/Pro \cite{gemini25_blog,gemini25_vertex}, Qwen2.5-Omni-3B and Qwen2.5-Omni-7B~\cite{qwen25_omni_2025}, Qwen2-Audio-7B-Instruct~\cite{qwen2_audio_2024}, Megrez-Omni-3B~\cite{li2025megrez}, and Step-Audio-2-mini~\cite{stepaudio2_2024}. These baselines span diverse architectures (dense/MoE), data scales, and training recipes, and collectively represent the state of the art in vision-language and speech-vision-text modeling in 2024–2025.To the best of our knowledge, there is no prior open-source, agriculture-specific omni-model; hence we compare AgriGPT-Omni primarily with strong general-purpose models under the same evaluation protocol.

We comprehensively evaluate AgriGPT-Omni across three major settings: text generation, vision-language generation, and multimodal speech understanding.

First, as shown in Table~\ref{tab:agribench13k-text}, AgriGPT-Omni outperforms all baselines on the AgriBench-13K text generation benchmark, surpassing strong LLMs such as Qwen2.5-VL and MiniCPM-V-2.6. This demonstrates that our domain-specific model retains strong textual reasoning capabilities, even in complex agricultural contexts.

Second, Table~\ref{tab:agribench-vl4k} reports results on AgriBench-VL-4K for image-text generation. AgriGPT-Omni again achieves the highest scores across all metrics, revealing its superior multimodal alignment and grounding ability, especially under agricultural visual distributions.

As shown in Tables~\ref{tab:Speech}--\ref{tab:a4-speechImageText}, AgriGPT-Omni consistently outperforms strong multimodal baselines across all four task types: speech open QA, speech–text multiple choice, speech–image open QA, and speech–image–text multiple choice. 

In speech-only and speech–image QA (Tables~\ref{tab:Speech} and \ref{tab:a3-speechImage}), our model achieves dominant pairwise win rates across all six languages, with particularly strong gains in dialectal scenarios such as Sichuanese and Cantonese. For multiple-choice settings (Tables~\ref{tab:speechtext} and \ref{tab:a4-speechImageText}), AgriGPT-Omni surpasses competing models by large margins—often by 10--20 accuracy points—regardless of whether visual information is included.

Overall, these results demonstrate that our three-stage training pipeline substantially enhances cross-lingual and cross-modal reasoning, yielding robust performance in complex audio-centric and multimodal agricultural tasks.

\begin{table*}[htbp]
\caption{Pairwise win rates of open QA (Q: Speech~$\rightarrow$~A: Text).}
\label{tab:Speech}
\centering
\small
\setlength{\tabcolsep}{12pt}
\begin{tabular}{lccccccc}
\toprule
Model & Cn & Kr & CnYue & En & CnSi & Ja & Win Rate \\
\midrule

Ours VS Qwen2.5-Omni-3B 
& 95.45\% 
& 99.73\% 
& 81.48\%  
& 64.71\% 
& 42.11\% 
& 94.12\% 
& 79.60\% \\

Ours VS Qwen2.5-Omni-7B 
& 50.00\% 
& 87.12\% 
& 76.47\% 
& 93.90\% 
& 75.00\% 
& 72.13\% 
& 75.77\% \\

Ours VS Megrez-Omni-3B 
& 100.00\% 
& 100.00\% 
& 95.31\% 
& 100.00\% 
& 96.97\% 
& 100.00\% 
& 98.74\% \\

Ours VS Qwen2-Audio-7B-Instruct 
& 98.89\% 
& 96.21\% 
& 94.44\% 
& 100.00\% 
& 98.81\% 
& 100.00\% 
& 97.85\% \\

Ours VS Step-Audio-2-mini 
& 100.00\% 
& 98.02\% 
& 96.61\% 
& 95.18\% 
& 92.06\% 
& 98.39\% 
& 96.71\% \\

\bottomrule
\end{tabular}
\end{table*}

\begin{table*}[htbp]
\caption{Accuracy on Multiple Choice (Q: Speech + Text ~$\rightarrow$~A: Text).}
\label{tab:speechtext}
\centering
\small
\setlength{\tabcolsep}{15pt}
\begin{tabular}{lccccccc}
\toprule
Model & Cn & Kr & CnYue & En & CnSi & Ja & ACC \\
\midrule

Qwen2.5-Omni-7B & 88.00\% & 77.00\% & 85.00\% & 87.00\% & 85.00\% & 81.00\% & 83.83\% \\
Step-Audio-2-mini & 84.00\% & 72.00\% & 83.00\% & 84.00\% & 81.00\% & 75.00\% & 80.83\% \\
Qwen2.5-Omni-3B & 86.00\% & 79.00\% & 84.00\% & 88.00\% & 83.00\% & 82.00\% & 83.67\% \\
Qwen2-Audio-7B-Instruct & 86.00\% & 80.00\% & 85.00\% & 80.00\% & 86.00\% & 73.00\% & 81.67\% \\
Megrez-Omni-3B & 80.00\% & 61.00\% & 75.00\% & 77.00\% & 79.00\% & 68.00\% & 73.33\% \\
Ours (AgriGPT-Omni) & 95.00\% & 91.00\% & 95.00\% & 92.00\% & 93.00\% & 90.00\% & 92.67\% \\

\bottomrule
\end{tabular}
\end{table*}

\begin{table*}[htbp]
\caption{Pairwise win rates of open VQA (Q: Speech + Image~$\rightarrow$~A: Text).}
\label{tab:a3-speechImage}
\centering
\small
\setlength{\tabcolsep}{13.5pt}
\begin{tabular}{lccccccc}
\toprule
Model & Cn & Kr & CnYue & En & CnSi & Ja & Win Rate \\
\midrule

Ours VS Qwen2.5-Omni-3B 
& 91.95\% 
& 98.05\% 
& 94.87\% 
& 91.67\% 
& 97.44\% 
& 95.12\% 
& 94.85\% \\

Ours VS Qwen2.5-Omni-7B 
& 89.89\% 
& 86.50\% 
& 95.89\% 
& 94.19\% 
& 92.31\% 
& 80.56\% 
& 89.89\% \\

Ours VS Megrez-Omni-3B 
& 92.75\% 
& 92.03\% 
& 92.96\% 
& 97.59\% 
& 91.43\% 
& 100.00\% 
& 94.46\% \\

\bottomrule
\end{tabular}
\end{table*}

\begin{table*}[htbp]
\caption{Pairwise win rates of Multiple Choice (Q: Speech + Image + Text~$\rightarrow$~A: Text).}
\label{tab:a4-speechImageText}
\centering
\small
\setlength{\tabcolsep}{16pt}
\begin{tabular}{lccccccc}
\toprule
Model & Cn & Kr & CnYue & En & CnSi & Ja & ACC \\
\midrule

Qwen2.5-Omni-7B & 63.00\% & 46.00\% & 61.00\% & 64.00\% & 60.00\% & 64.00\% & 59.17\% \\

Qwen2.5-Omni-3B & 64.00\% & 54.00\% & 59.00\% & 68.00\% & 57.00\% & 56.00\% & 59.67\% \\

Megrez-Omni-3B & 49.00\% & 50.00\% & 45.00\% & 62.00\% & 43.00\% & 56.00\% & 50.83\% \\

AgriGPT-Omni & 78.00\% & 72.00\% & 76.00\% & 80.00\% & 76.00\% & 71.00\% & 75.67\% \\

\bottomrule
\end{tabular}
\end{table*}

\begin{table}[htbp]
\centering
\small
\setlength{\tabcolsep}{5pt}
\caption{Accuracy comparison between Base model(Qwen2.5-Omni-7B) and AgriGPT-Omni-7B.}
\label{tab:generality}
\begin{tabular}{lcc}
\toprule
Dataset & Qwen2.5-Omni-7B & AgriGPT-Omni-7B \\
\midrule
MMLU (Text)             & 49.97\% & 62.81\% \\
ARC (Text)              & 81.69\% & 82.19\% \\
OpenBookQA (Text)       & 60.34\% & 80.74\% \\
MMBench (Vision)        & 83.45\% & 82.51\% \\
MMMU (Vision)           & 44.64\% & 43.75\% \\
SeedBench (Vision)      & 74.87\% & 71.92\% \\
Sample Audio (Audio)    & 75.57\% & 74.49\% \\
\bottomrule
\end{tabular}
\end{table}

\subsection{Ablation study}

As shown in Figure~\ref{fig:ablation}, we evaluate stepwise performance changes across four representative task types: Open QA (Speech → Text), Open QAA (Speech + Image → Text), Multiple Choice (Speech + Text → Text), and Multiple Choice (Speech + Image + Text → Text). We adopt Win Rate for open-ended generation tasks and Accuracy for multiple-choice tasks to provide a comprehensive view of model behavior under different reasoning constraints.

Across all tasks, we observe consistent, monotonic, and interpretable improvements from the base model through Phase 1, Phase 2, and Phase 3. Phase 1 introduces large-scale text knowledge, which substantially enhances linguistic semantics, factual grounding, and instruction-following capabilities. Phase 2 performs multimodal alignment, enabling the model to fuse visual and auditory cues more effectively and improving its ability to perform cross-modal reasoning in complex agricultural scenarios. Phase 3 applies GRPO reinforcement learning, delivering the most significant final gains—particularly on multiple-choice evaluations—by refining decision boundaries, eliminating inconsistent outputs, and stabilizing multimodal preference patterns.

These trends highlight clear functional complementarities among the three training stages. Text knowledge injection equips the model with a strong language core, multimodal alignment provides structural coherence across modalities, and reinforcement learning yields robust behavior under ambiguous or noisy inputs. The cumulative effect of these stages results in substantial improvements in robustness, cross-modal consistency, and generalization, allowing AgriGPT-Omni to reliably handle the full modality spectrum encountered in real-world agricultural workflows. The steady upward trajectory across all task types further confirms that our training pipeline yields predictable and controllable model enhancements, which is essential for safe and trustworthy deployment in practical settings.

\subsection{Generalization Evaluation}

To assess whether domain-specialized training harms general capability, we evaluate AgriGPT-Omni on a suite of widely used benchmarks covering text reasoning (MMLU~\citep{hendrycks2021measuring}, ARC~\citep{clark2018think}, OpenBookQA~\citep{mihaylov2018can}), vision understanding (MMBench~\citep{liu2023mmbench}, MMMU~\citep{yue2024mmmu}, SeedBench~\citep{liu2024seedbench}), and audio comprehension (Sample Audio~\citep{chan2021speechstew}). AgriGPT-Omni achieves accuracy comparable to the base Qwen2.5-Omni-7B model across all modalities, and even surpasses it on several text benchmarks (e.g., +12.8 on MMLU and +20.4 on OpenBookQA). These results demonstrate that our targeted multimodal fine-tuning enhances agricultural capability while preserving—and in some cases improving—general-domain performance, indicating strong generalization and no overfitting to the agricultural domain.

\subsection{Real-world robustness evaluation.} 
As shown in Tables~\ref{tab:synthetic-vs-human}, to assess whether our model can reliably operate in real-world environments, we conduct a direct comparison between synthetic speech and human-recorded speech inputs—a critical requirement for deployment in settings where audio quality, recording devices, and speaker conditions vary widely. As shown in Table~\ref{tab:synthetic-vs-human}, the model achieves 234 wins, 111 ties, and 232 losses, demonstrating that its performance generalizes well beyond curated training conditions. The comparable outcomes between synthetic and human speech indicate that the model is not overly dependent on controlled or studio-quality inputs, and can maintain stable reasoning and comprehension even when exposed to noisy, spontaneous, or heterogeneous speech sources.

Where real deployments often involve low-resource communities, field data, and non-expert users. The ability to handle both synthetic and real human audio without degradation suggests that our system is suitable for scalable, accessible, and inclusive AI services, supporting practical use cases such as agricultural advisories, crisis-response communication, multilingual information access, and other high-stakes applications where reliable speech understanding is essential.

\section{Conclusion}

We present \textbf{AgriGPT-Omni}, the first agricultural omni-modal system that unifies speech, vision, and text understanding within a single framework. To support real-world agricultural voice–image–text interactions, we develop a scalable data construction pipeline that synthesizes and collects the largest multilingual agricultural speech dataset to date, combining 490K high-quality TTS samples with 1.4K human-recorded training speech and an additional 586 human-recorded samples for real-world evaluation. Building on this dataset, we train an agriculture-domain omni-model through a three-stage paradigm of textual knowledge injection, progressive multimodal alignment, and GRPO-based reinforcement learning.  

To enable rigorous evaluation, we introduce the first full-modality agricultural benchmark covering four speech-centric task formats and six languages. Experiments demonstrate that AgriGPT-Omni substantially improves speech and multimodal reasoning in agricultural scenarios while maintaining strong general-domain capability across text, vision, and audio benchmarks.

Overall, AgriGPT-Omni establishes a comprehensive foundation for speech-driven agricultural intelligence and opens new possibilities for field diagnostics, farmer assistance, and multimodal human–AI interaction. All datasets, models, and evaluation tools will be released to support future research and real-world deployment.

\section{Appendices}

\subsection*{A.1\quad Group Relative Policy Optimization (GRPO)}

\subsubsection*{A.1.1\quad GRPO objective}

During GRPO training, for each iteration and a given input $q$, we sample
$M$ candidate outputs $\{y_j\}_{j=1}^M$ from the reference policy
$\pi_{\mathrm{ref}}$. Each candidate $j$ receives a reward $r_j$.
We compute the group-relative advantage as
\begin{equation}
  \tilde{A}_j \;=\; \frac{r_j - \nu}{\tau}, \qquad
  \nu \;=\; \frac{1}{M}\sum_{j=1}^{M} r_j,\qquad
  \tau \;=\; \sqrt{\frac{1}{M}\sum_{j=1}^{M} (r_j-\nu)^2}.
  \label{eq:grpo-adv}
\end{equation}
Here, $\nu$ and $\tau$ are the mean and standard deviation of rewards
within the group.

Let the importance ratio be
\begin{equation}
  \varrho_j \;=\; \frac{\pi_\phi(y_j \mid q)}{\pi_{\mathrm{ref}}(y_j \mid q)}.
  \label{eq:ratio}
\end{equation}
The clipped surrogate objective of GRPO is

\begin{equation}
\begin{aligned}
  \mathcal{L}_{\text{GRPO}}(\phi)
  = \mathbb{E}_{y_j \sim \pi_{\mathrm{ref}}}\!\Bigg[
    \frac{1}{M}\sum_{j=1}^{M}
    \min\!\Big(
      \varrho_j \tilde{A}_j,\;
      \operatorname{clip}(\varrho_j,\,1-\epsilon,\,1+\epsilon)\,\tilde{A}_j
    \Big)
  \Bigg]
  \\
  \hspace{3.5cm}
  - \lambda\,\mathrm{KL}\!\left[\pi_\phi \,\|\, \pi_{\mathrm{ref}}\right].
\end{aligned}
\label{eq:grpo-obj}
\end{equation}

\subsubsection*{A.1.2\quad Reward design and implementation}

We adopt a simple, deterministic reward for multiple-choice or
short-answer tasks. Given a model completion $c$ and the ground-truth
answer $a^\star$, we first extract a candidate answer $\hat{a}$:
\begin{equation}
  \hat{a} \;=\;
  \begin{cases}
    \text{the substring between }\texttt{\textbar\textbar}\,\cdot\,\texttt{\textbar\textbar},
    & \text{if present},\\[4pt]
    \text{the trimmed completion } c,
    & \text{otherwise}.
  \end{cases}
  \label{eq:extract}
\end{equation}

The per-sample reward is an exact-match score with a positive margin:
\begin{equation}
  r(c,a^\star) \;=\;
  \begin{cases}
    2.0, & \hat{a}=a^\star,\\
    0.0, & \text{otherwise}.
  \end{cases}
  \label{eq:reward}
\end{equation}

For a group of $M$ candidates $\{y_j\}_{j=1}^M$ sampled from
$\pi_{\mathrm{ref}}$, we compute rewards $\{r_j\}_{j=1}^M$ via
\eqref{eq:reward}, then normalize them to obtain the group-relative
advantages $\{\tilde{A}_j\}_{j=1}^M$ using \eqref{eq:grpo-adv}. These
advantages enter the clipped surrogate objective in \eqref{eq:grpo-obj}
together with the importance ratios in \eqref{eq:ratio} and the KL term.

\subsubsection*{A.1.3\quad Multilingual edit-distance reward (WER/CER)}

For speech or transcription-style tasks, we employ a language-aware
edit-distance reward. Given a completion $c$ and a reference transcript
$a^\star$, we compute an error rate $\mathrm{err}(c,a^\star)$ as

\begin{equation}
  \mathrm{err}(c,a^\star) \;=\;
  \begin{cases}
    \mathrm{WER}(a^\star,c), & \text{if language is English (``en'')},\\[4pt]
    \mathrm{CER}(a^\star,c), & \text{otherwise}.
  \end{cases}
  \label{eq:wercer-select}
\end{equation}

Here WER is the word error rate and CER is the character error rate.
Their standard definitions are
\begin{equation}
  \mathrm{WER}(a^\star,c)
  \;=\;
  \frac{S + D + I}{N_{\text{w}}},\qquad
  \mathrm{CER}(a^\star,c)
  \;=\;
  \frac{S + D + I}{N_{\text{ch}}},
  \label{eq:wercer-def}
\end{equation}
where $S,D,I$ denote the numbers of substitutions, deletions and
insertions w.r.t.\ $a^\star$, and $N_{\text{w}}$ and $N_{\text{ch}}$
are the word and character counts of $a^\star$, respectively.

We map the error to a bounded, positive reward via a linear transform:
\begin{equation}
  r(c,a^\star)
  \;=\;
  2\bigl(1 - \mathrm{err}(c,a^\star)\bigr),
  \qquad
  r(c,a^\star) \leftarrow \max\!\bigl(0,\, r(c,a^\star)\bigr).
  \label{eq:wercer-reward}
\end{equation}
Thus $r\!\in\![0,2]$ whenever $\mathrm{err}\!\in\![0,1]$. The lower
clipping improves robustness when an implementation returns values
slightly above $1$ due to tokenization or normalization mismatches.

For each prompt, a group of $M$ candidates
$\{y_j\}_{j=1}^M$ is scored by \eqref{eq:wercer-reward} to produce
$\{r_j\}_{j=1}^M$, which are then normalized into group-relative
advantages $\{\tilde{A}_j\}_{j=1}^M$ using \eqref{eq:grpo-adv}. The
advantages feed the GRPO objective \eqref{eq:grpo-obj} together with
the importance ratios in \eqref{eq:ratio}.

\bibliographystyle{ACM-Reference-Format}
\bibliography{sample-base}

\appendix

\end{document}